\def\year{2019}\relax
\documentclass[letterpaper]{article} 
\usepackage{aaai19}  
\usepackage{times}  
\usepackage{helvet}  
\usepackage{courier}  
\usepackage{url}  
\usepackage{graphicx}  
\usepackage{booktabs}       
\usepackage{amsfonts}       
\usepackage{nicefrac}       
\usepackage{microtype}      
\usepackage{graphicx}

\usepackage{algorithm}
\usepackage{algpseudocode}
\usepackage{amsmath}
\usepackage{mathrsfs}

\usepackage{subcaption}
\usepackage{enumerate}

\begin{document}
\title{KTAN: Knowledge Transfer Adversarial Network}
\author{
Peiye Liu$^1$, 
Wu Liu$^2$, 
Huadong Ma$^1$,
Tao Mei$^2$, 
Mingoo Seok$^3$,
\\
$^1$ Beijing University of Posts and Telecommunications,Beijing, China\\
$^2$ JD AI Research, Beijing, China\\
$^3$ Department of Electrical Engineering Columbia University New York, New York, U.S.A  \\
\{liupeiye, mhd\}@bupt.edu.cn,
\{liuwu1,tmei\}@jd.com,
ms4415@columbia.edu
}
\maketitle

\begin{abstract}
	To reduce the large computation and storage cost of a deep convolutional neural network, the knowledge distillation based methods have pioneered to transfer the generalization ability of a large (teacher) deep network to a light-weight (student) network.
	However, these methods mostly focus on transferring the probability distribution of the softmax layer in a teacher network and thus neglect the intermediate representations.
	In this paper, we propose a knowledge transfer adversarial network to better train a student network. Our technique holistically considers both intermediate representations and probability distributions of a teacher network.
	To transfer the knowledge of intermediate representations, we set high-level teacher feature maps as a target, toward which the student feature maps are trained.
	Specifically, we arrange a Teacher-to-Student layer for enabling our framework suitable for various student structures.
	The intermediate representation helps the student network better understand the transferred generalization as compared to the probability distribution only.
	Furthermore, we infuse an adversarial learning process by employing a discriminator network, which can fully exploit the spatial correlation of feature maps in training a student network.
	The experimental results demonstrate that the proposed method can significantly 
	improve the performance of a student network on both image classification and object detection tasks.
\end{abstract}

\section{Introduction}

The AlexNet \cite{krizhevsky2012imagenet} and various other deep convolutional neural network (CNN) models have demonstrated the state-of-the-art performance in computer vision tasks such as image classification \cite{he2016deep}, object detection \cite{girshick2015fast}, and pose estimation \cite{liu2017weighted}. However, the top-performance CNN models generally employ very wide and deep architecture consisting of a numerous number of synapses and neurons \cite{simonyan2014very}. Training and deploying such complex CNN models indeed incur large computation and storage cost, which limits the implementation of a CNN on a resource-limited device.
To tackle the challenges, researchers have attempted techniques to accelerate the computation of CNN models. These techniques can be roughly divided into three types: network quantization \cite{courbariaux2016binarized,zhou2017incremental}, network pruning \cite{han2015deep,li2017pruning}, and knowledge distillation (KD) \cite{hinton2015distilling,luo2016face}. Network quantization methods attempt to convert a pre-trained full-precision CNN model into a low-precision one \cite{zhou2017incremental,cheng2017survey}. Network pruning methods attempt to remove the redundant and insignificant connections (weights) \cite{han2015deep}.

On the other hand, KD methods aim to train a light-weight model with the knowledge transferred from a large model that is trained. For example, Hinton \emph{et al.}~\cite{hinton2015distilling} collects the outputs of the softmax layer (probability distribution) of a teacher network and use them as target objectives in training the student network. Despite its simplicity, KD demonstrates promising results in several classification tasks \cite{hinton2015distilling}.
However, if we consider extracting the final probability distribution as the knowledge to transfer, its application can be limited to only the classification tasks with the softmax loss function.

To avoid such problem, recent studies~\cite{romero2014fitnets,zagoruyko2016paying} proposed to exploit intermediate representations as sharable knowledge. Specifically, they use the outputs in the convolutional layers of a teacher network.
As a high-dimensional feature distribution, the knowledge in feature maps consists of the feature values and their spatial correlations, which are requisite in various deep CNN models. 
For transferring the shareable knowledge, they directly align the values of intermediate representations of the teacher and student network.
Admittedly, it works for the transferring of the probability distribution. However, for the intermediate representation, such direct aligning cannot effectively transfer the latent spatial correlation. Given the importance of such information in computer vision tasks, the direct aligning remains as a critical limitation. It also ignores the distinction between the distribution spaces of the teacher and the student networks since their topological differences would make them generalize with the different distributions.

In this paper, we aim to address the aforementioned challenges. We propose a new framework that is based on a knowledge transfer adversarial network (KTAN).
The knowledge transfer (KT) process, which is a general class of the KD method, is divided into two parts: 1) knowledge extraction and 2) knowledge learning processes.
In the first \textit{knowledge extraction} step, since the deeper convolutional layer extracts more complicated and high dimensional features, we choose the feature maps of teacher's last convolutional layer as the shared knowledge which contains pixel-level as well as spatial information.
Most, if not all, of CNN architectures contain the convolutional layers, which enable our framework applicable to the networks that have no softmax layer and thus cannot use the existing KD method. In the second \textit{knowledge learning} step, we adopt the concept of the Generative Adversarial Networks (GAN) and propose to employ three networks in the knowledge transfer adversarial framework: 1) a teacher generative network (TGN); 2) a student generative network (SGN); and 3) a discriminator network (DN). The TGN observes a large network model and generates the teacher feature map (TFM) as shared knowledge. The SGN is a light-weight network model. The TGN and SGN are firstly trained on the ground truth for initialization, respectively. Considering different sizes and channels between TFM and SFM, we present a Teacher-to-Student layer in TGN to match the size of the student feature map (SFM).
After well trained, we exploit the MSE loss in SGN to learn the similar SFM with TFM. In the adversarial training stage, the optimization target of DN tries to understand the spatial information in the shared knowledge through distinguishing whether an output came from TGN rather than SGN. Differently, SGN attempts to learn the spatial information through maximizing the probability of being classified as SFM by the discriminator. Besides, the entire student network, including convolutional layers and fully connected layers, is also optimized by the original task with ground truth. The illustration of our framework is presented in Figure 1.

In the end, we propose a knowledge transfer adversarial teacher-student framework for various student networks. In addition, because of the utilization of the feature maps shared knowledge, our framework is suitable for various computer vision tasks, such as classification and detection. The evaluations on image classification and object detection benchmarks demonstrate that our method certainly improves the performance of different student networks.

To summarize, the contributions of this work are as follows:

\begin{itemize}
\item{We propose a knowledge transfer adversarial network to endow the light-weight student network training with more affluent intermediate representation knowledge from a deeper teacher model;}
\item{We extend the teacher-student framework with Teacher-to-Student layer for arbitrary structures of student network and deliver the spatial information in the shared knowledge to a student network in an adversarial learning manner;}
\item{Extensive experiments conducted on both image classification and object detection tasks verify the merit of our knowledge transfer adversarial network (KTAN)}
\end{itemize}

\begin{figure*}[t]
    \begin{center}
        \includegraphics[width=.9\textwidth]{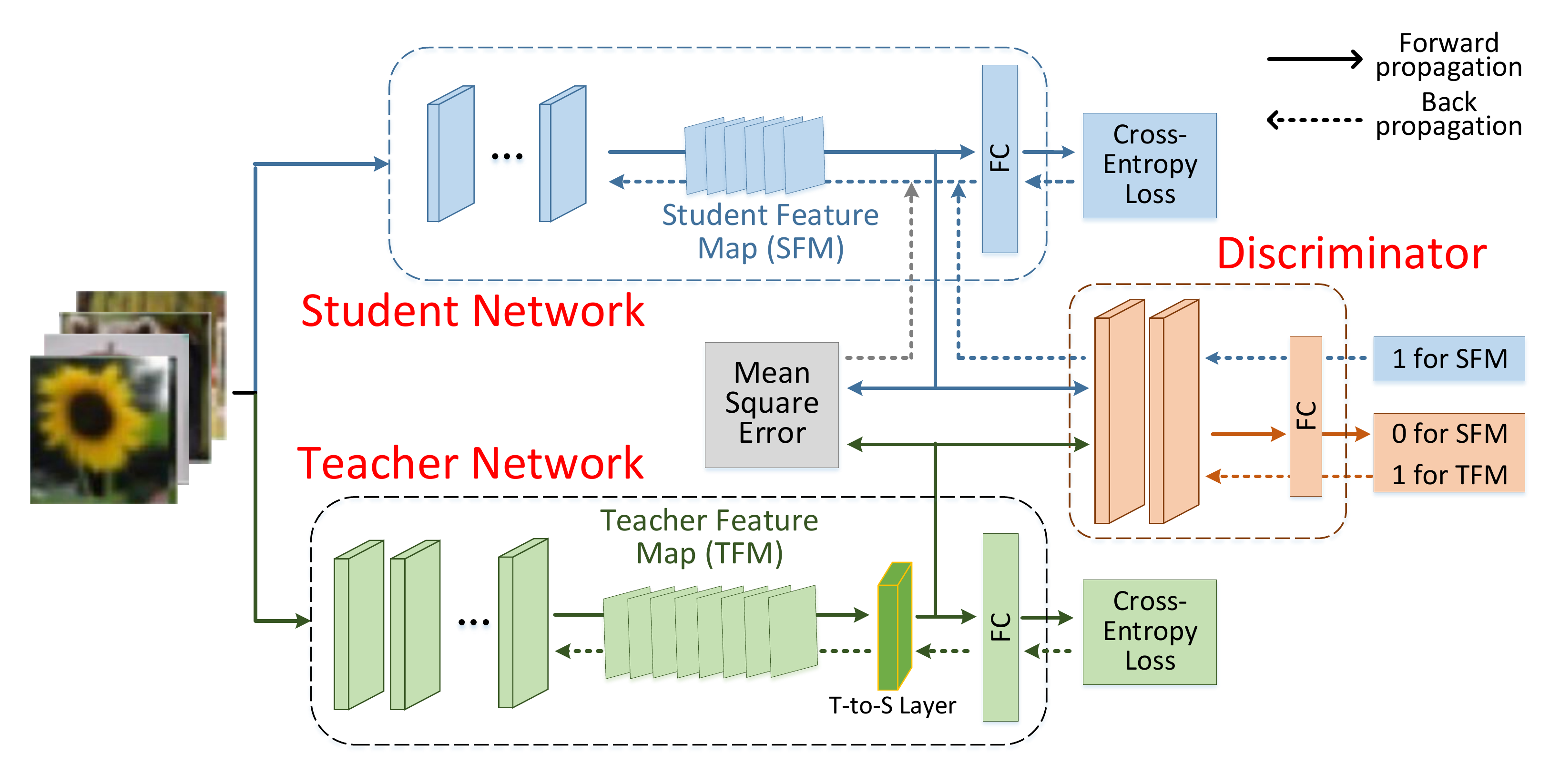}
        \caption{The architecture of the Knowledge Transfer Adversarial Network on classification task. The green lines represent Feedfoward (FW) and Backpropagation (BP) of the teacher network, the blue lines represent FW and BP of the student network, and the orange lines represent FW and BP of the discriminator network. (Best seen in color)}
        \label{fig:fig1}
    \end{center}
\end{figure*}

\section{Related Works}

As mentioned above, the KT issue studied in this paper focus on transferring the generalization ability of a large teacher model to a small student model. If the teacher model performs well, a student model trained to generalize in the same way will typically achieve better results than a small model trained in the normal way. Therefore, there are two targets in KT issue, including how to extract the shared knowledge from a large teacher model and how to transfer it to a simpler student model.

Among the early works on KT, \cite{hinton2015distilling} concluded the softmax output extracted from a large teacher model involves the information about the large model's way to distinguish the correct and wrong classes. If a teacher model generalizes well because of complicated structure, a small network trained to generalize in the same way will typically achieve much better results. Nevertheless, the softmax output can only provide the information about the classification ability of a teacher model and relies on the number of classes. Other computer vision tasks optimize different objective function, for example, detection problem adopts bounding boxes of instances as the target objective. Therefore, the softmax output cannot be applied to other computer vision tasks.

After that, researchers attempted to extract intermediate representation from a teacher model for optimizing the knowledge extraction process. \cite{romero2014fitnets} extended the idea of KD and introduced FitNet to compress a network from wide and relatively shallow to thin and deep. In order to learn the generalization of a teacher network, FitNet adopted a squared difference objective function to make the student models mimic the middle layer output of the teacher network. Although the middle layer output in CNN models offer the knowledge to explain how a teacher network generalize, the directly matching learning solution ignored the correlation in it. Therefore, a student model can hardly learn the teacher model’s way to generalize. Later, \cite{zagoruyko2016paying} proposed an idea of Attention Transfer (AT). This method encoded the spatial areas of a teacher network mostly focusing by attention maps and transferred the attention maps as shared knowledge to a student network. However, this method also utilized a directly matching learning function as the objective function to train a student model. Although the attention maps may contain useful spatial information generalized by the teacher network, the directly matching learning process could not sufficiently transfer the spatial generalization of a teacher network to a student network.

\section{Knowledge Transfer Adversarial Network}

For better performance, CNN models merge more filters and convolutional layers to form a deeper and wider network structure \cite{he2016deep}. On the contrary, in some realistic scene, the model is required to be pruned or quantized for satisfying the demand of the resource-limited devices \cite{howard2017mobilenets}. Considering the hardness of the trade-off between those two targets, we provide a method to transfer the knowledge from a large teacher network to a small student network. This section is divided into four parts to introduce our teacher-student knowledge transfer framework. Sec 3.1 introduces the process of extracting the shared knowledge from a large teacher model. Sec 3.2 presents a normal method for transferring the shared knowledge from a teacher network to student network. Considering the different structure between two models, we design a teacher-to-student regressor layer for matching the size of TFM and SFM in Sec 3.3. In order to make up the deficiency of the directly matching learning method, we propose an adversarial knowledge learning framework for understanding the spatial information in shared knowledge in Sec 3.4.

\subsection{Knowledge Extraction}

Computer vision is the academic field which aims to gain a high-level understanding of the low-level information given by raw pixels from digital images. Deep CNN based approaches have been exposed great achievement on computer vision tasks, such as classification \cite{geifman2017selective}, localization \cite{wei2017deep}, detection \cite{newell2017associative} and segmentation \cite{hu2017maskrnn}.

In a CNN model, it trains multiple convolutional layers to extract different feature from simple to complex. Different from fully connected (FC) layers, each convolutional layer aims to train multiple linear image filters for capturing a more complicated visual feature from output of last layer. The filter $F\in\mathbb{R}^{k_{w}\times k_{h}\times c}$ is convolved with the multiple channel of input images or feature maps $I\in\mathbb{R}^{w\times h\times c}$ from last layer to produce a new image $I'$.


\begin{equation}
\begin{split}
I'(x,y)=\sum_{i_{x}=1-\lceil\frac{k_w}{2}\rceil}^{\lfloor\frac{k_w}{2}\rfloor}\sum_{i_{x}=1-\lceil\frac{k_h}{2}\rceil}^{\lfloor\frac{k_h}{2}\rfloor}\sum_{i_c=1}^{c} &I(x+i_x,y+i_y,i_c)\\
&\cdot F(i_x,i_y,i_c)
\end{split}
\end{equation}

where $k_w$ and $k_h$ represent the kernel width and height of $F$, $w$ and $h$ represent the size of input images or feature maps, $c$ represents the number of channel of input images.

As proved by \cite{zeiler2014visualizing}, a trained shallow convolutional layer shows responds on low-level features, like edge, angle, and curve. Then, the next convolutional layer generates responds to a more complicated feature, like circle and rectangle. Therefore, as we ascend the layers, the convolutional layer extracts a more complicated and high dimensional feature. On the other hand, a deep convolutional feature also represents the generalization ability of the network better than the shallow one. In this case, we choose the feature maps (FM) of the last convolutional layer of teacher network as the shared knowledge, which consists of pixel level value information and spatial information.

\subsection{Knowledge Directly Learning}

After we obtain the shared knowledge from the teacher network, an effective transfer function is required to guide a student network to learn the generalization of the teacher network. An obvious way to transfer the shared knowledge is encouraging a student network to simulate the output of a teacher network. In this method, a Mean Square Error (MSE) is adopted as extra objective function for training the student network. Considering the FM of the student network as $m_{s}\in\mathbb{R}^{w\times h\times c}$ and the FM of the teacher network as $m_{t}\in\mathbb{R}^{w'\times h'\times c'}$, an extra loss function can be calculated by:

\begin{equation}
L_{MSE}=\frac{1}{r^2cwh}\sum_{n=1}^{c}\sum_{x=1}^{rw}\sum_{y=1}^{rh}(m_{t}(x,y,n)-m_{s}(x,y,n))^2
\end{equation}
Where $r$ is the scale ratio, $w$ and $h$ are the width and height of $m_{s}$.

As shown in Equation 2, the MSE objective function aims to train a student network by aligning the pixel value of SFM and TFM. However, because the shared knowledge consists of pixel values and correlation spatial information between pixels and channels, this method ignores the spatial information in the shared knowledge.

\begin{algorithm}[htb!]
	\caption{Training process of the Teacher-to-Student layer.}
	\begin{algorithmic}[1]
		\Require
		The Teacher network model $T$, divided to convolutional parts $G$ and fully connected parts $C$;
		\Ensure
		The weight of Teacher-to-Student layer, $w_r$;
		\State Load pretrained $G$, $C$, and initial the Teacher-to-Student layer $w_r$ randomly;
		\For{number of k step training iterations}
		\State Input $N$ training samples $\{I^{1},...,I^{N}\}$ with label $\{y^{1},...,y^{N}\}$ to the teacher network.
		\State Update the $w_r$ by cross-entropy loss $\mathcal{H}=(\textbf{y},C(R(G(i))))$
		\EndFor
	\end{algorithmic}
\end{algorithm}

\subsection{Knowledge Regressor}

As mentioned above, large neural networks become wider and deeper for achieving better performance \cite{he2016deep}. While small neural networks cut redundant synapses and layers for deploying in resource-limited devices \cite{howard2017mobilenets}. Hence, SFM and TFM usually have different size, which blocks the knowledge learning process. Therefore, we add a teacher-to-student regressor to the end of the last convolutional layer in the teacher network, whose output matches the size of the SFM. Considering keeping the spatial information in the shared knowledge and less memory consumption, we define a convolutional regressor layer for resizing the high dimensional feature maps. Let $M_{t, w}\times M_{t, h} $ and $C_{t}$ be the TFM's spatial size and number of channels. Correspondingly, let $M_{l, w}\times M_{l, h} $ and $C_{l}$ be the SFM's spatial size and number of channels. Given a shared knowledge of size $( C_{t}, M_{t, w}\times M_{t, h})$, the teacher-to-student regressor sets the output channel as $C_{l}$ of learning layer and adopts its kernel size by $(M_{t, i}+2\times P -K_{i})/S_{i}+1=M_{l, i}$, where $i\in \{h, w\}$. In the training stage, given the teacher network $T$, the detailed training process of Teacher-to-Student layer is shown in Algorithm 1.
Therefore, we obtain a regressed shared knowledge with the same size of student network.

\subsection{Knowledge Adversarial Learning}

For the general GAN model, two parallel networks, generative network (GN) and discriminator network (DN), are trained alternatively to improve each other \cite{goodfellow2014generative}. GN learns to generate the real data distribution, and DN distinguishes whether a sample came from the groundtruth rather than GN. Next, the tightly intertwined adversarial training process can significantly improve the performance of GN and DN. For the KT problem, we propose a KTAN framework which aims to transfer a shared knowledge from a teacher network into a small network by the adversarial learning process, as shown in Figure 1.

Different from the original GAN, our KTAN contains three networks, a teacher generative network (TGN), a student generative network (SGN), and a discriminator network (DN). TGN adopts a large network model and takes an image $I$ as the input to generate a TFM as shared knowledge. SGN adopts a simpler network model and takes the same input image $I$ as the input to generate a SFM. Considering the TFM's class discrimination property, DN employs a shallow VGG-like network which contains only one convolutional layer. Because of the CNN structure, DN can understand the spatial information in TFM and map it into a probability distribution space $q$. Therefore, if we map SFM into the same probability distribution space $q$ by DN, the student network can be trained through the cross-entropy loss of DN for generating a similar SFM with TFM.

In the adversarial training process, to obtain the teacher network's TFM distribution $m_{t}$ over image data $i$, we represent a mapping to data space as $T({i};w_{t})$, the green actual lines in Figure 1, where $T$ is the convolutional network parts of the teacher network with wight $w_{t}$.
Then, a Teacher-to-Student layer is defined by $R$ with weights $w_{r}$ for regressing the output of teacher network and represents a mapping to feature map space as $R(m_{t};w_{r})$, green actual lines in Figure 1.
To transfer the spatial information in TFM, we feed $m_{s}$ and $R(m_{t})$ to discriminator model $D$, the actual orange lines in Figure 1, to train $D$ to maximize the probability of identifying the correct label to both feature maps from Teacher-to-Student layer and the student network $S$.
The student network is simultaneously trained to minimize the distinction between $D(m_{s})$ and $D(R(m_{t}))$ through $log(1-D(S({i})))$, original objective function and MSE difference between $m_{s}$ and $R(m_{t})$.
In the training stage of the whole framework, we firstly pretrain the layer $R$ with image data $i$ for obtaining the $w_{r}$, as presented in Algorithm 1. Then, considering the feature maps extracted from the teacher network are high-level abstract information and easily to classification, which leads a low probability of $D$ making a mistake, we devise $k$ steps of optimizing $S$ with the original task's loss function and MSE object function before adversarial optimizing. After that, we simultaneously train the $S$ and $D$ playing the following two-player minimax game and a detailed example training process on classification task is presented in Algorithm 2.

\begin{algorithm}[htb!]
	\caption{Training process of KTAN on classification task.}
	\label{alg:process}
	\begin{algorithmic}
		\Require
		The Student network model, divided to generator $S$ part and classifier $C$ part;
		The trained Teacher-to-Student layer $R$ and Teacher network $T$;
		\Ensure
		The improved student model, $S$ and $C$;
		\State Load pretrained S, C, T and initial the discriminator network D randomly;
		Pretrain the Teacher-to-Student layer R;
		\For{number of k step pretraining iterations}
		\State Input n training samples ${I^1,...,I^n}$ with label ${y^1,...,y^n}$ to Student networks.
		\State Update the $S$ by cross-entropy loss LCE and the Mean square error loss between $m_s$ and $R(m_t)$
		\begin{equation*}
		L_{CE}(y,C(m_s))+\beta L_{MSE}(R(m_t),m_s)
		\end{equation*}
		\EndFor
		\For {number of adversarial training iterations}
		\State Input same $N$ training samples ${I^{1},...,I^{N}}$ to $G$ and $T$.
		\State Sample $N$ student feature maps $m_{s}$ from $S$.
		\State Sample $N$ teacher feature maps $R(m_{t})$ from $R$.
		\State \textbf{Update} network $D$ by $L_{D}$, i.e., the probability that $m_{s}$ is considered as a teacher network's generalization:
		\begin{equation*}
			L_{D}=\sum_{n-1}^{N}-logD(m_{s})
		\end{equation*}
		\State \textbf{Update} network $S$ by:
		\begin{equation*}
				\mathcal{H}(\textbf{y},C(m_{s}))+\alpha L_{D}(y_{R(m_{t})},m_{s})+\beta L_{MSE}(R(m_{t})),m_{s})
		\end{equation*}
		\State \textbf{Update} network $C$ by:
		\begin{equation*}
			\mathcal{H}(\textbf{y},C(m_{s}))
		\end{equation*}
		\EndFor
	\end{algorithmic}
\end{algorithm}

\section{Experiment}

In this section, we conduct two computer vision tasks to verify our knowledge transfer adversarial network model. As the most popular and traditional issue in computer vision, image classification problem is selected as our first task to show the performance of our method on knowledge transfer. Besides, considering benefits of our model, we extend extra experiments on object detection problems to demonstrate the generalization ability of our model.

\subsection{Image classification}

For classification problems, we evaluate our model on two standard datasets, CIFAR-10 and CIFAR-100. The CIFAR is a popular image classification benchmark. It contains 50k training images and 10K testing images with 10 and 100 classes, where instances are 32 x 32 color images involving airplanes, cats, human and so on. In the experiments, we utilize the random horizontal flips and random crops data argumentation. For general training, SGD method with a mini-batch size of 32 is selected to optimize the training process beginning with learning rate 0.2. For the adversarial training parts, we set the learning rate starting from $10^{-2}$ and the weight decay to $10^{-4}$.

\begin{table}[htb!]
	\centering
	\caption{Knowledge transfer results on CIFAR}
	\begin{tabular}{lcc}
		\hline
		\textbf{Method} & \textbf{CIFAR10(\%)} & \textbf{CIFAR100(\%)} \\
		\hline
		Student & 93.69 & 73.10  \\
		KD [Hinton et al.] & 94.70  & 76.05  \\
		FitNet [Romero et al.] & 94.44  & 75.26  \\
		AT [Zagoruyko et al.] & 94.53  &  74.56 \\
		DLN & 94.47  &  75.30 \\
		KTAN & 94.73  &  75.67 \\
		\textbf{KD+KTAN} & \textbf{94.91}  &  \textbf{76.34} \\
		\hline
		Teacher & 95.12  & 78.03  \\
		\hline
	\end{tabular}
\end{table}

On the CIFAR datasets, we choose a very deep residual network Resnet-1001 \cite{he2016identity} as the teacher network and a shallow version of Inception network \cite{ioffe2015batch} as the student network. Further, we compare our model with several state-of-the-art knowledge transfer methods, including KD \cite{hinton2015distilling}, FitNet \cite{romero2014fitnets} and AT \cite{zagoruyko2016paying}.

\begin{enumerate}[(1)]
	\item \textbf{Teacher}. A large CNN model (Resnet-1001) trained by the true label objective, which contains 1001 layers.
	\item \textbf{Student}. A small CNN model (Inception) trained by the true label objective.
	\item \textbf{KD} \cite{hinton2015distilling}. This method utilizes the softmax output of a teacher network as shared knowledge. We raise the softmax temperature for teacher network to 4, and set the weight given to the teacher cross-entropy to 0.9, following \cite{hinton2015distilling}.
	\item \textbf{FitNet} \cite{romero2014fitnets}. This method utilizes an intermediate representation as shared knowledge and applying a direct knowledge learning process. Considering we train a simpler student network instead of a thin and deep one, which needs more regularization from the teacher network, we transfer the last convolutional layer's output to a student network. The weight given to the transfer loss is four, following \cite{romero2014fitnets}.
	\item \textbf{AT} \cite{zagoruyko2016paying}. Only utilizing the attention maps as shared knowledge and applying a direct knowledge learning process. Considering the different structure of Teacher and Student, we can only align the attention maps of the last convolutional layer in two networks. The weight given to the transfer loss is 0.05, following the explanation in \cite{zagoruyko2016paying}.
	\item \textbf{Directly learning network (DLN)}. Our KTAN network without the adversarial learning process.
	\item 	\textbf{KTAN}. Utilizing the FM of a deep convolutional layer and applying an adversarial knowledge learning process. We set the $\alpha$ to 0.6 and $\beta$ to 0.5.
	\item \textbf{KTAN + KD}. We combine our KTAN and KD to transfer both the FM and softmax output to \textbf{student}. The adversarial learning process is only applied on FM shared knowledge.
\end{enumerate}

As shown in Table 1, our KTAN model indeed improves the performance of the original student network, which indicates the effectiveness of the intermediate representation based adversarial learning process. Comparing with other methods, our KTAN model is also competitive. In the CIFAR-10 dataset, KTAN archives the best performance among the methods. DLN method performs a little better than FitNet. The reason is the regressor layer used in FitNet will hold some parts of the shared knowledge. For AT method, because of different structures between the teacher and student networks, it is hard to map attention maps of all convolutional layers from a teacher to a student network. In certain circumstances, only the last convolutional layer’s attention map is suitable for transferring to a student network, which contains few spatial information than a high-level generalization of a deep convolutional layer. After that, our KTAN method shows better performance than DLN, which indicates the adversarial training process in the KTAN model indeed help a student network understand the spatial information better. In the CIFAR-100 dataset, since more classes provide more information about a large model's way to classify on the probability distribution, the softened softmax output achieves better results than the one on CIFAR-10. Combining with it, our KTAN model also achieves the best performance on CIFAR-100.

\begin{table}[htb!]
	\centering
	\caption{mAP result on Pascal VOC 2007 dataset}
	\begin{tabular}{lcc}
		\hline
		\textbf{Method} & \textbf{Architecture} & \textbf{mAP} \\
		\hline
		Student & Faster-RCNN (Res50) & 70.43  \\
		KD & Faster-RCNN (Res50)  &  70.92 \\
		FitNet & Faster-RCNN (Res50)  &  71.41 \\
		DLN & Faster-RCNN (Res50)  &  71.54 \\
		\textbf{KTAN} & Faster-RCNN (Res50)  & \textbf{72.78} \\
		\hline
		\textbf{KTAN+KD} & Faster-RCNN (Res50)  & \textbf{72.91} \\
		\hline
		Teacher & Faster-RCNN (Res152)  & 75.47  \\
		\hline
	\end{tabular}
\end{table}

\begin{figure}[t!]
	\begin{subfigure}[b]{.33\linewidth}
		\centering
		\includegraphics[width=.9\textwidth]{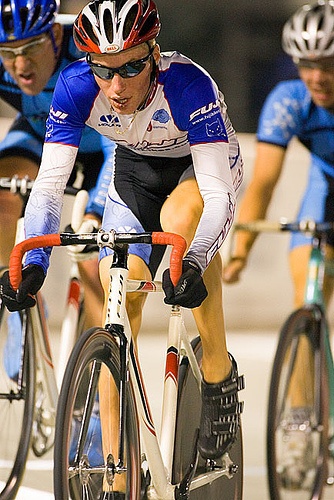}
	\end{subfigure}%
	\begin{subfigure}[b]{.33\linewidth}
		\centering
		\includegraphics[width=.9\textwidth]{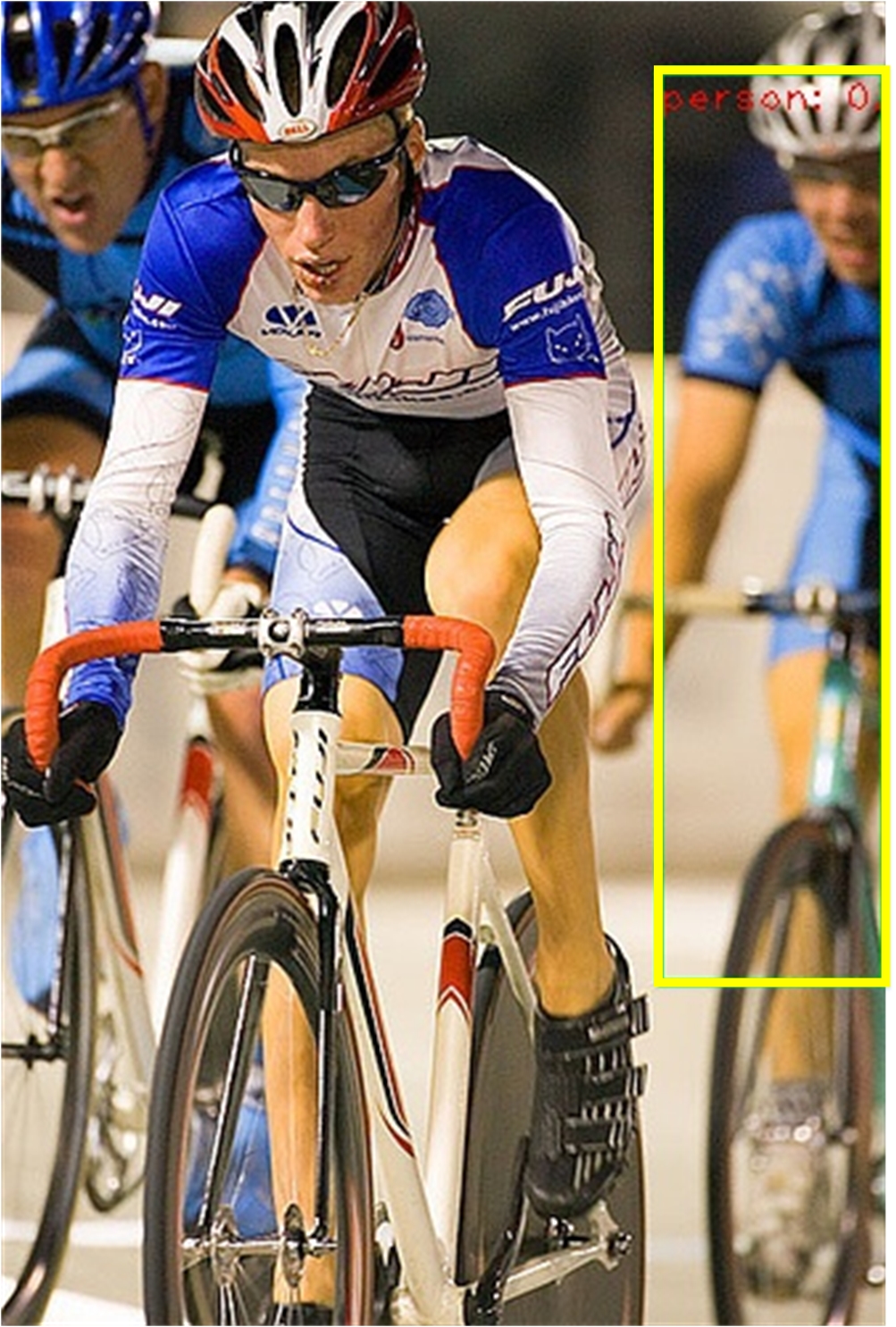}
	\end{subfigure}%
	\begin{subfigure}[b]{.33\linewidth}
		\centering
		\includegraphics[width=.9\textwidth]{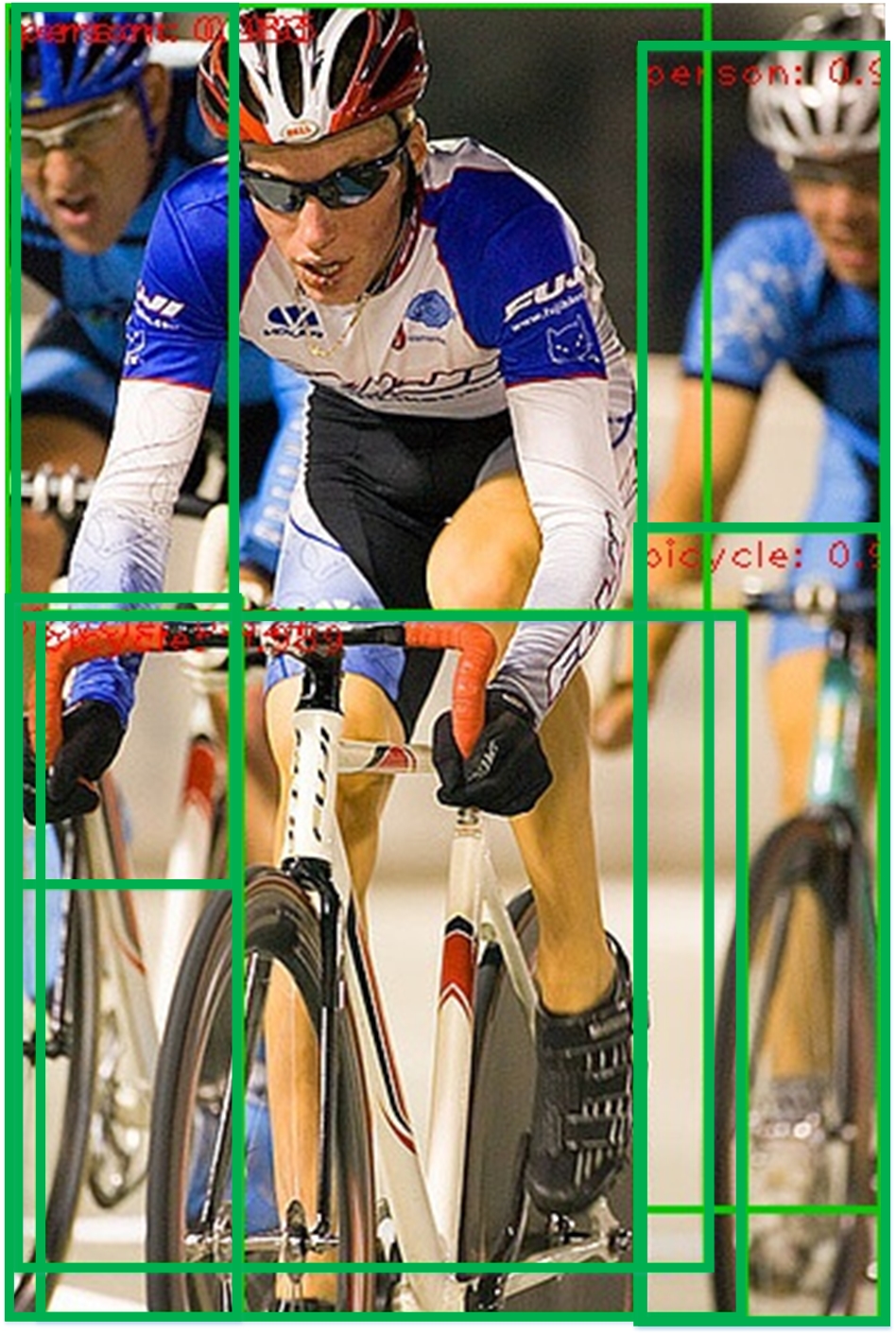}
	\end{subfigure}%
	\vspace{3pt}\\
	\begin{subfigure}[b]{.33\linewidth}
		\centering
		\includegraphics[width=.9\textwidth]{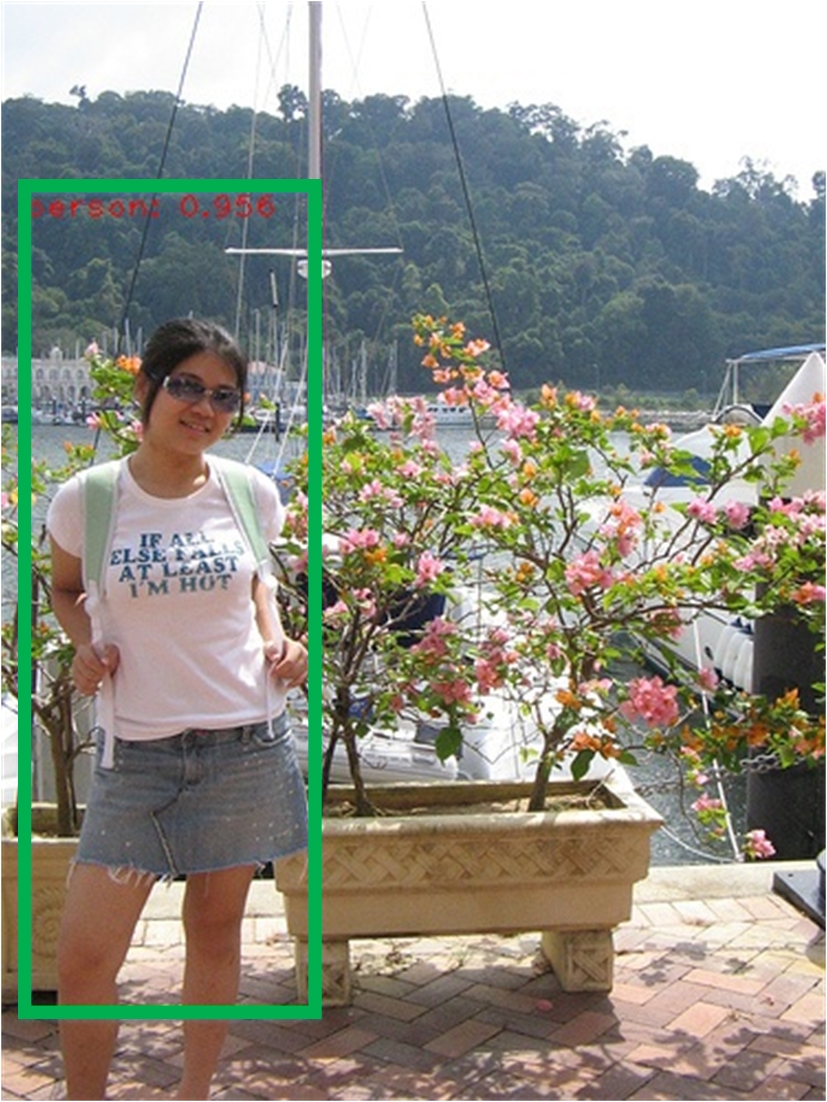}
	\end{subfigure}%
	\begin{subfigure}[b]{.33\linewidth}
		\centering
		\includegraphics[width=.9\textwidth]{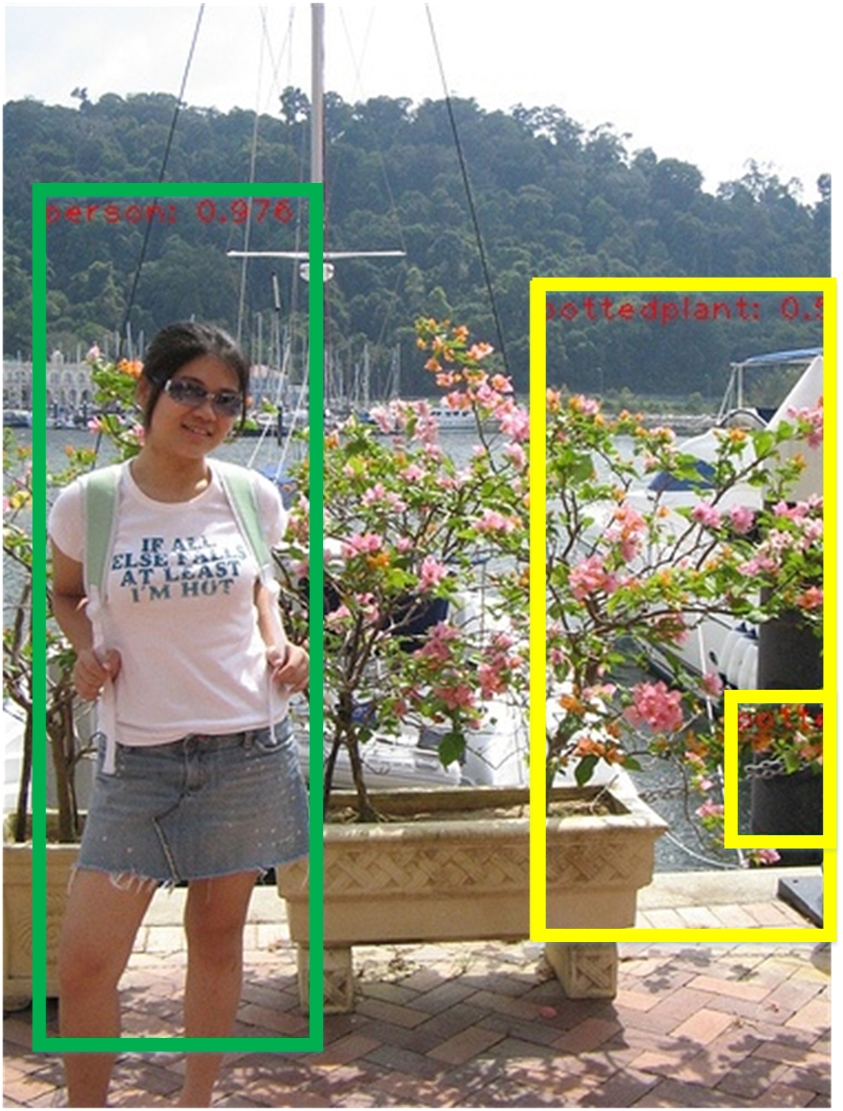}
	\end{subfigure}%
	\begin{subfigure}[b]{.33\linewidth}
		\centering
		\includegraphics[width=.9\textwidth]{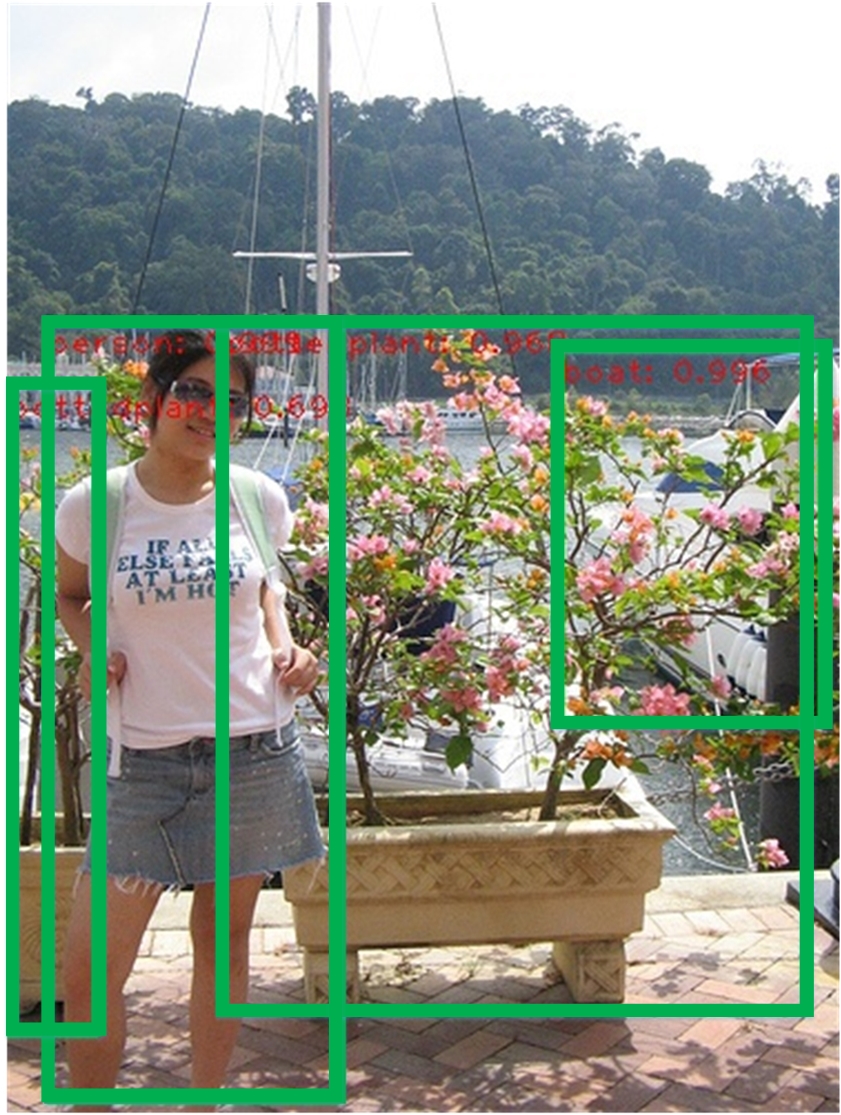}
	\end{subfigure}%
	\vspace{3pt}\\
	\begin{subfigure}[b]{.33\linewidth}
		\centering
		\includegraphics[width=.9\textwidth]{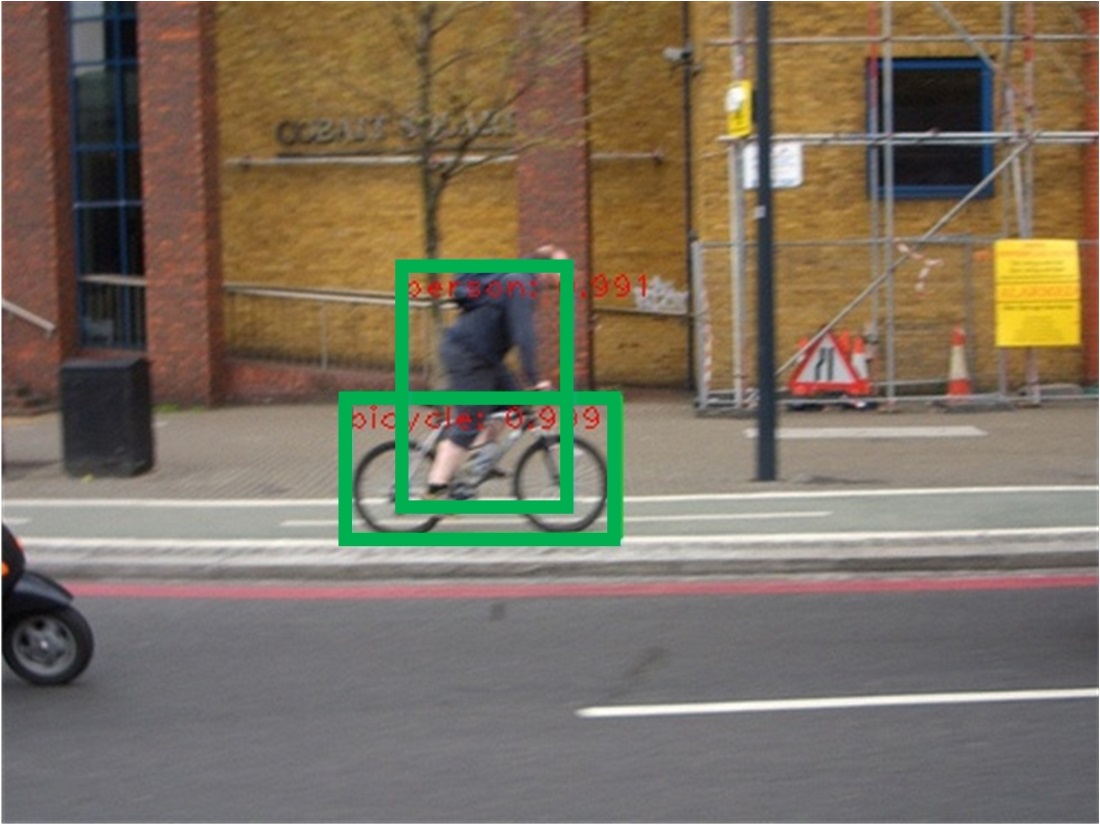}
	\end{subfigure}%
	\begin{subfigure}[b]{.33\linewidth}
		\centering
		\includegraphics[width=.9\textwidth]{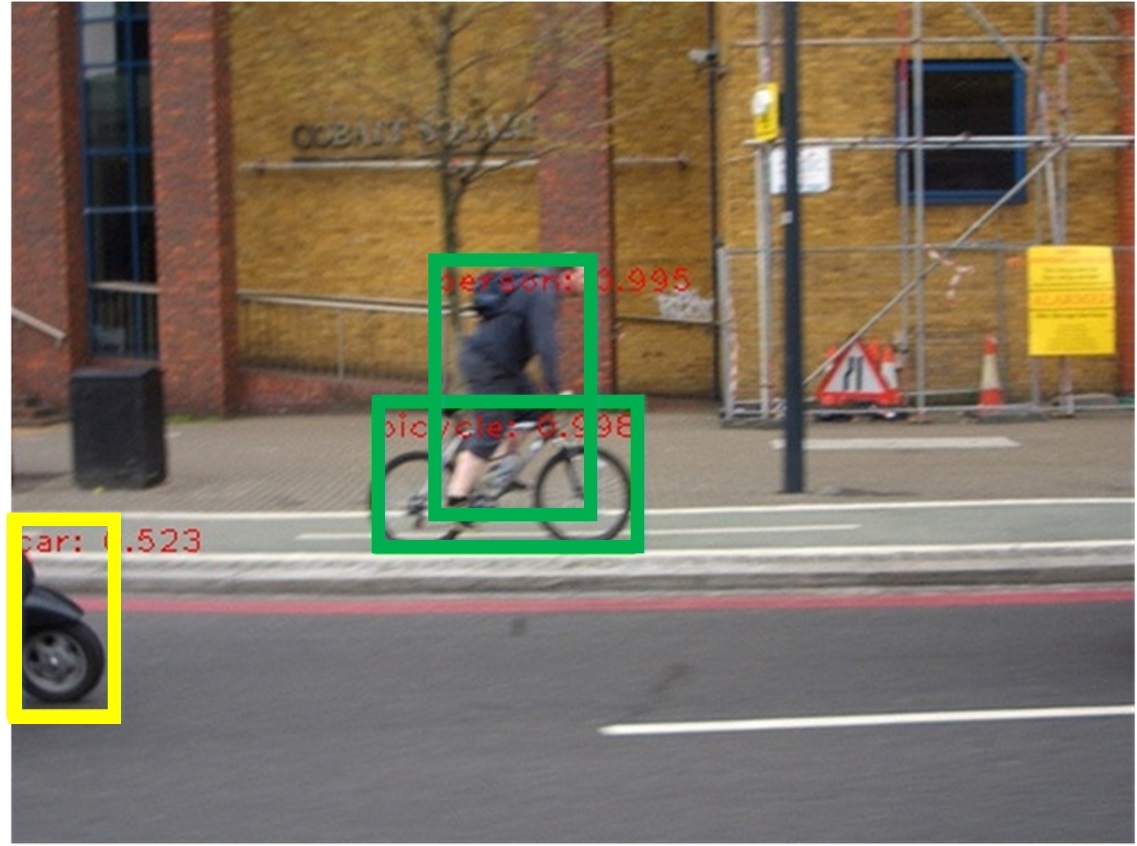}
	\end{subfigure}%
	\begin{subfigure}[b]{.33\linewidth}
		\centering
		\includegraphics[width=.9\textwidth]{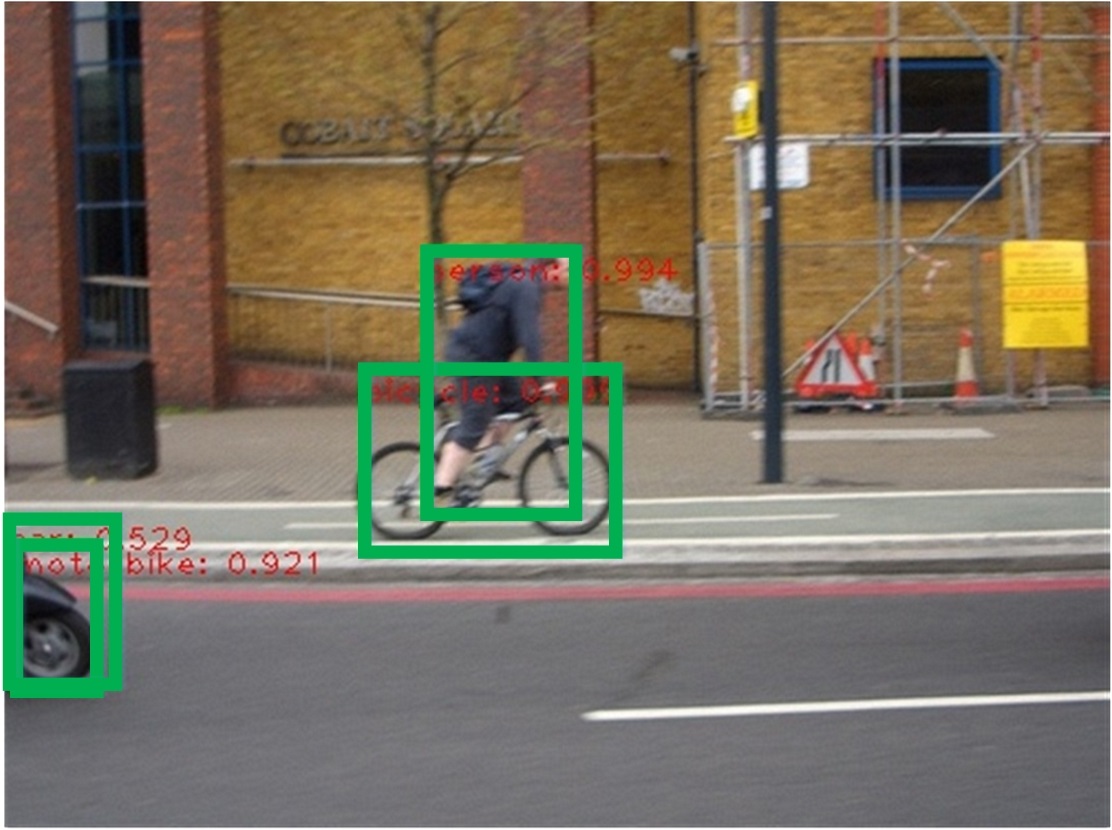}
	\end{subfigure}%
	\vspace{3pt}\\
	\begin{subfigure}[b]{.33\linewidth}
		\centering
		\includegraphics[width=.9\textwidth]{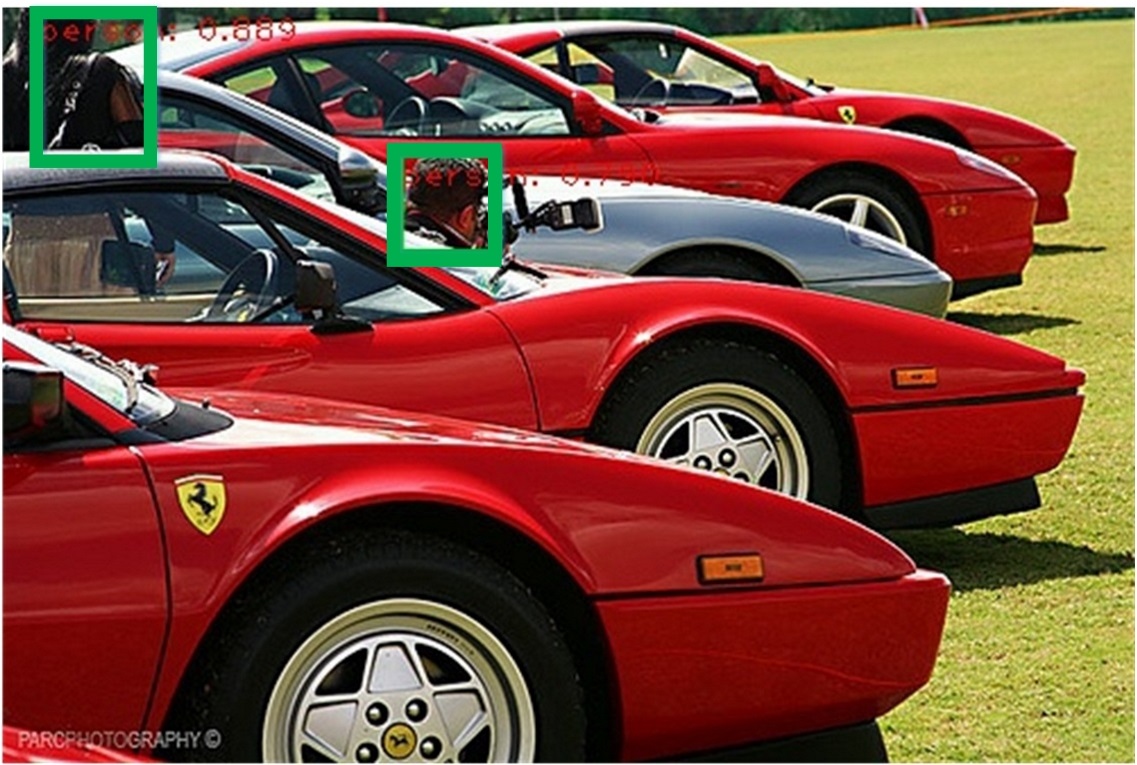}
	\end{subfigure}%
	\begin{subfigure}[b]{.33\linewidth}
		\centering
		\includegraphics[width=.9\textwidth]{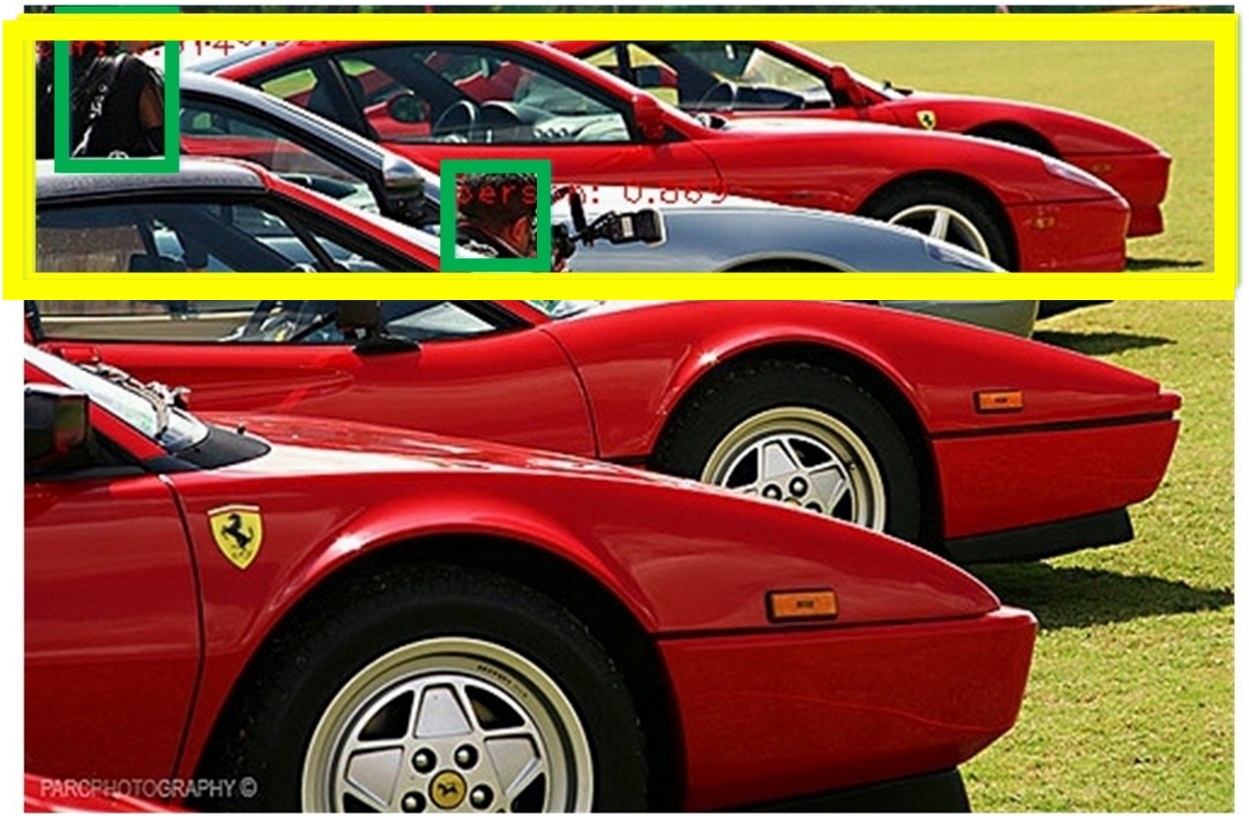}
	\end{subfigure}%
	\begin{subfigure}[b]{.33\linewidth}
		\centering
		\includegraphics[width=.9\textwidth]{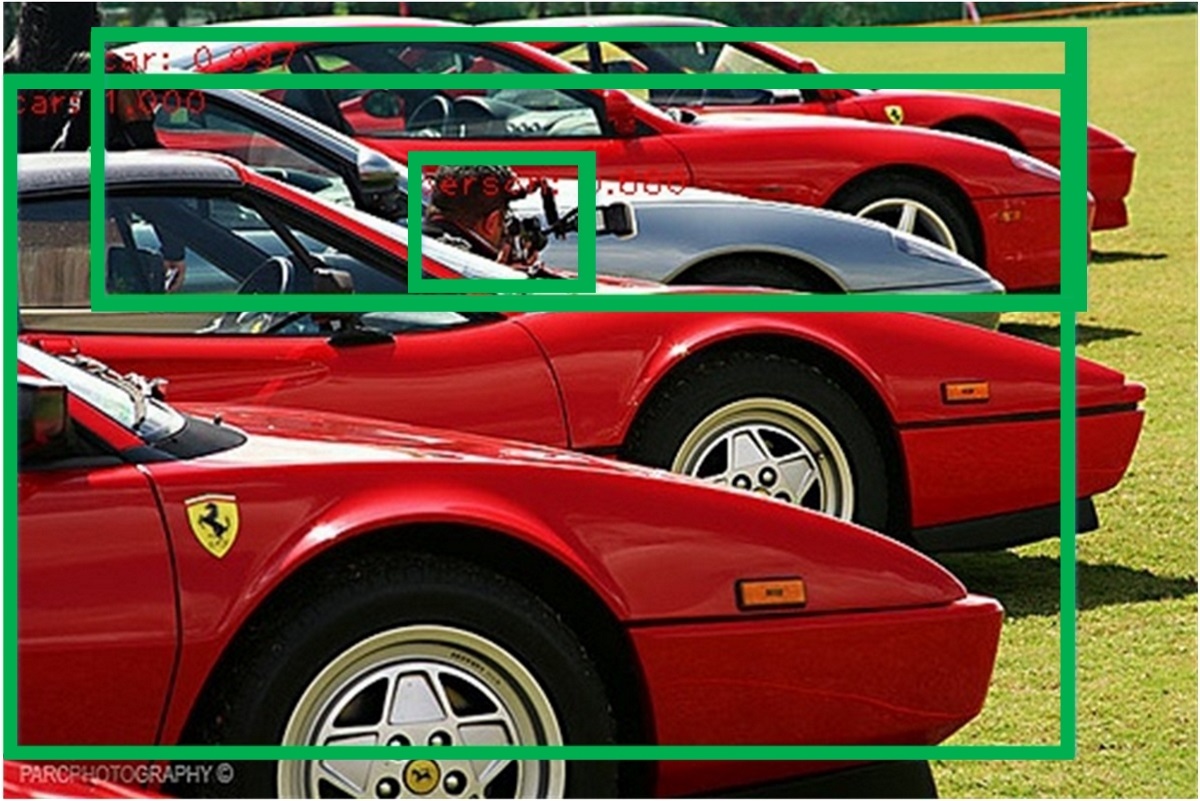}
	\end{subfigure}%
	\vspace{3pt}\\
	\begin{subfigure}[b]{.33\linewidth}
		\centering
		\includegraphics[width=.9\textwidth]{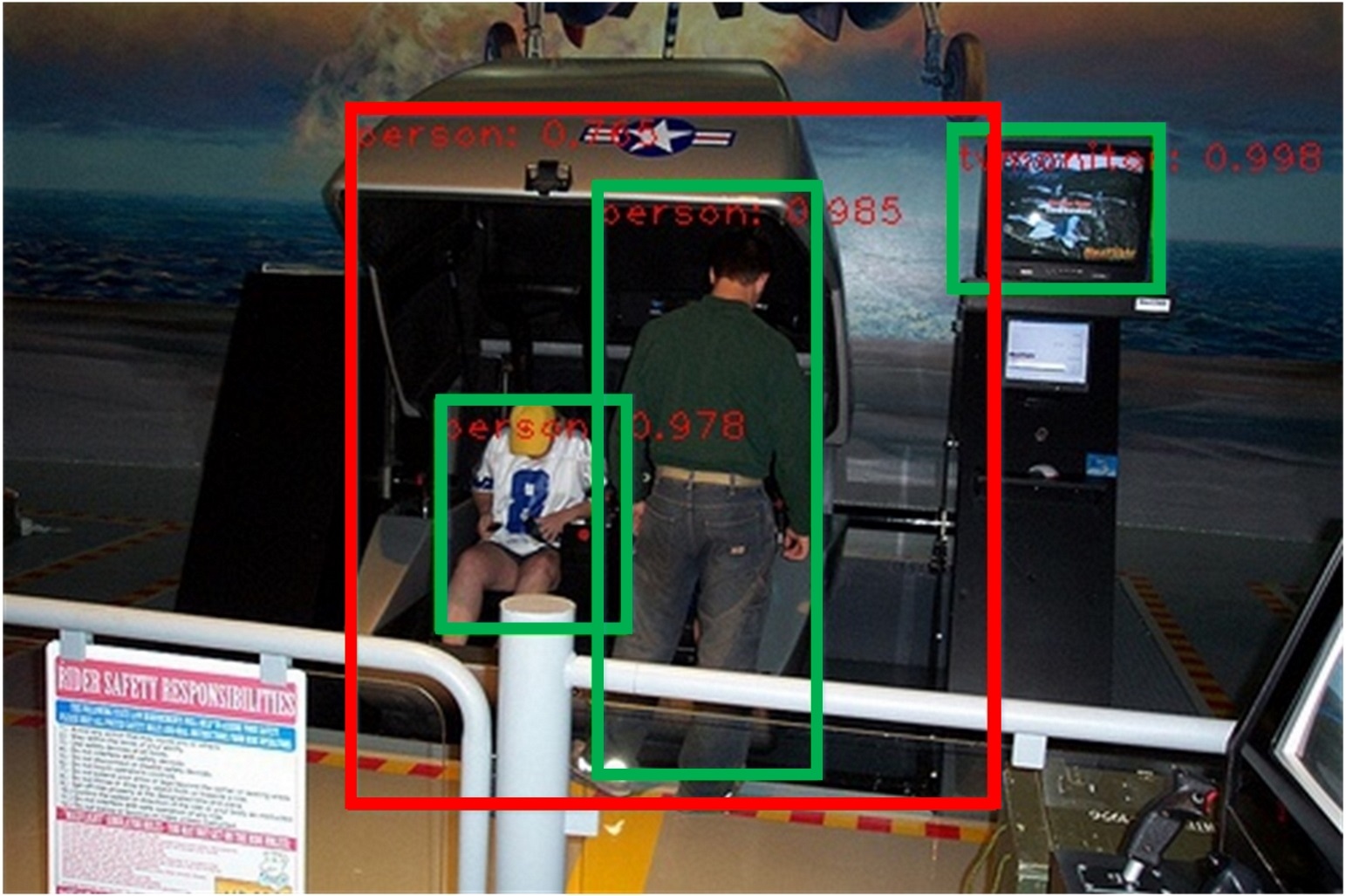}
	\end{subfigure}%
	\begin{subfigure}[b]{.33\linewidth}
		\centering
		\includegraphics[width=.9\textwidth]{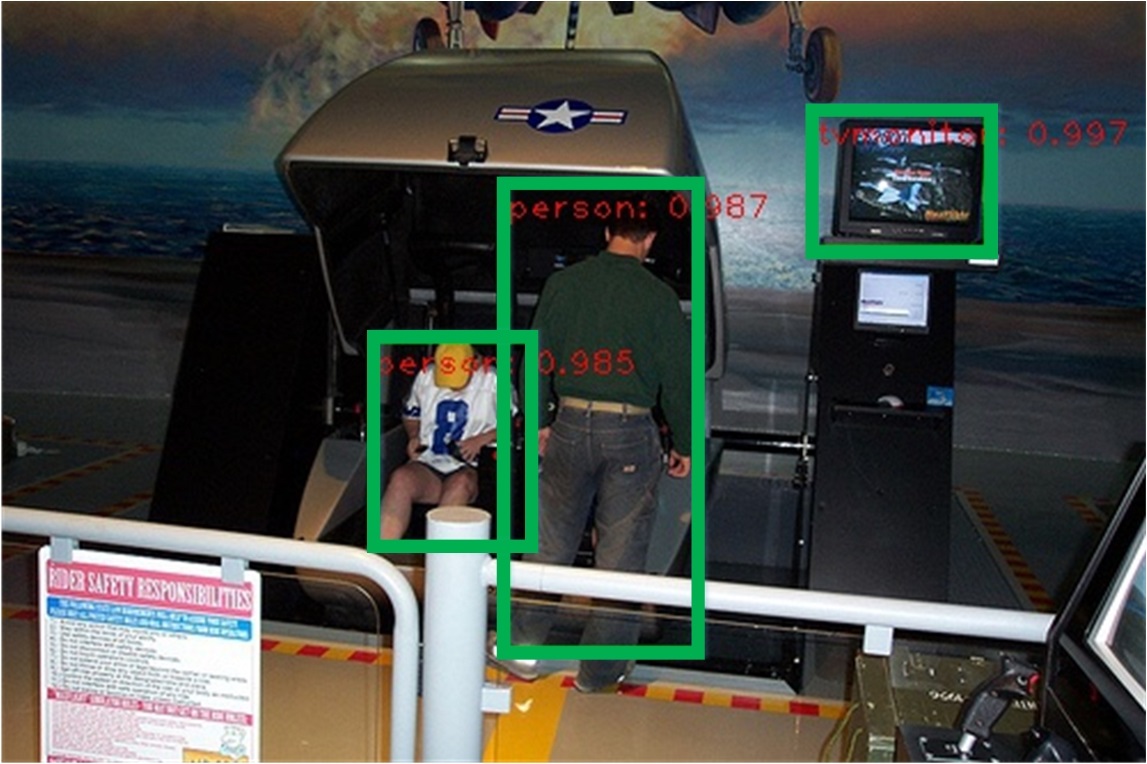}
	\end{subfigure}%
	\begin{subfigure}[b]{.33\linewidth}
		\centering
		\includegraphics[width=.9\textwidth]{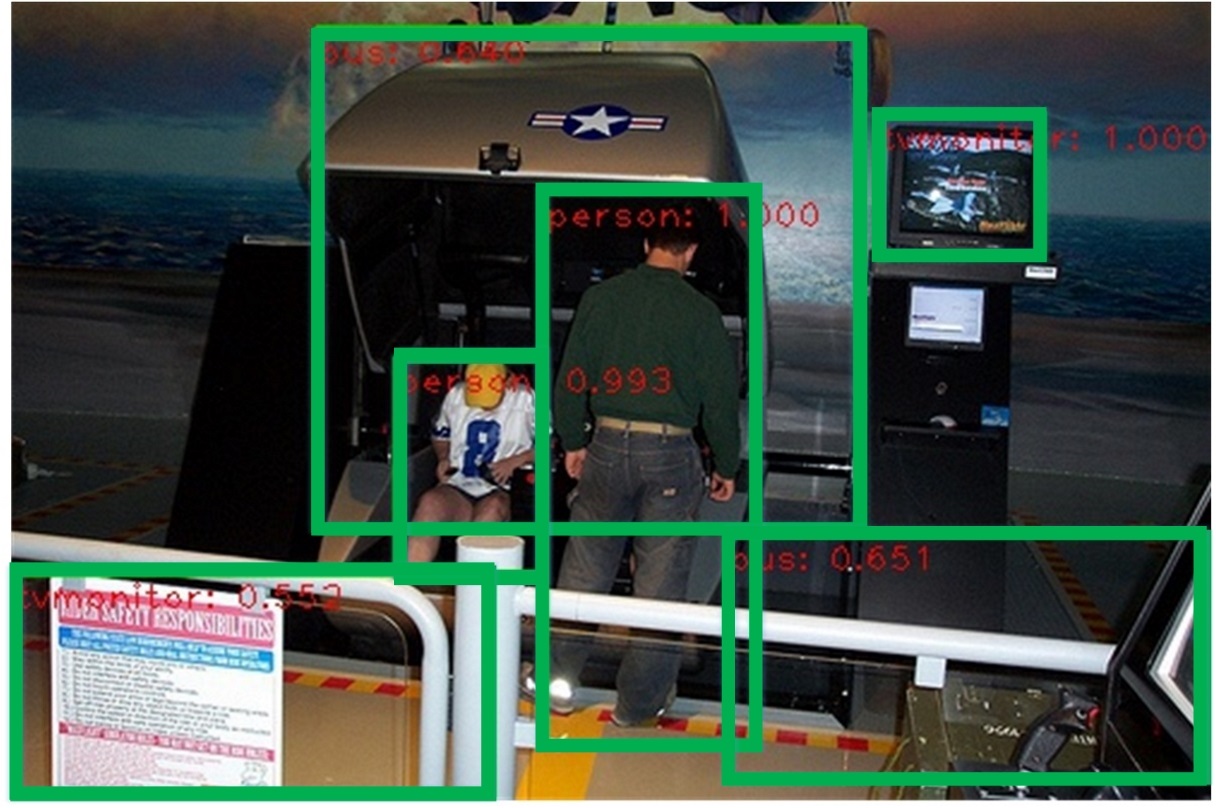}
	\end{subfigure}%
	\vspace{3pt}\\
	\begin{subfigure}[b]{.33\linewidth}
		\centering
		\includegraphics[width=.9\textwidth]{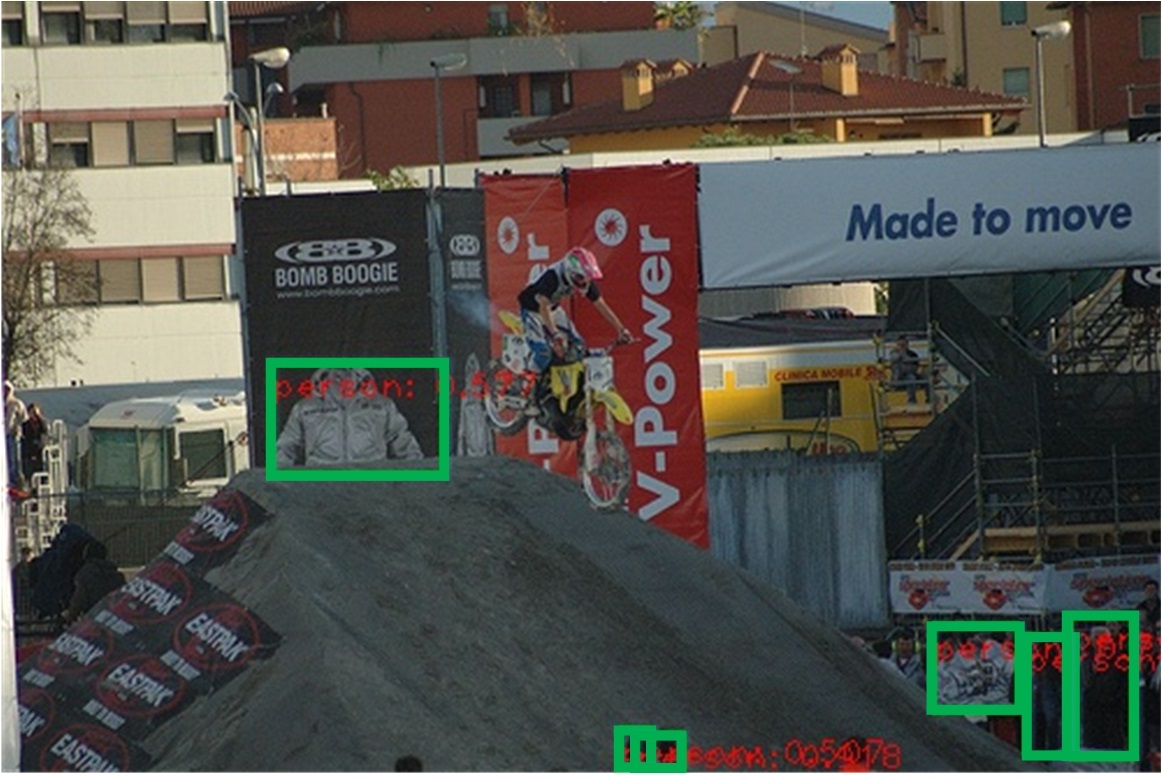}
		\caption{ }
	\end{subfigure}%
	\begin{subfigure}[b]{.33\linewidth}
		\centering
		\includegraphics[width=.9\textwidth]{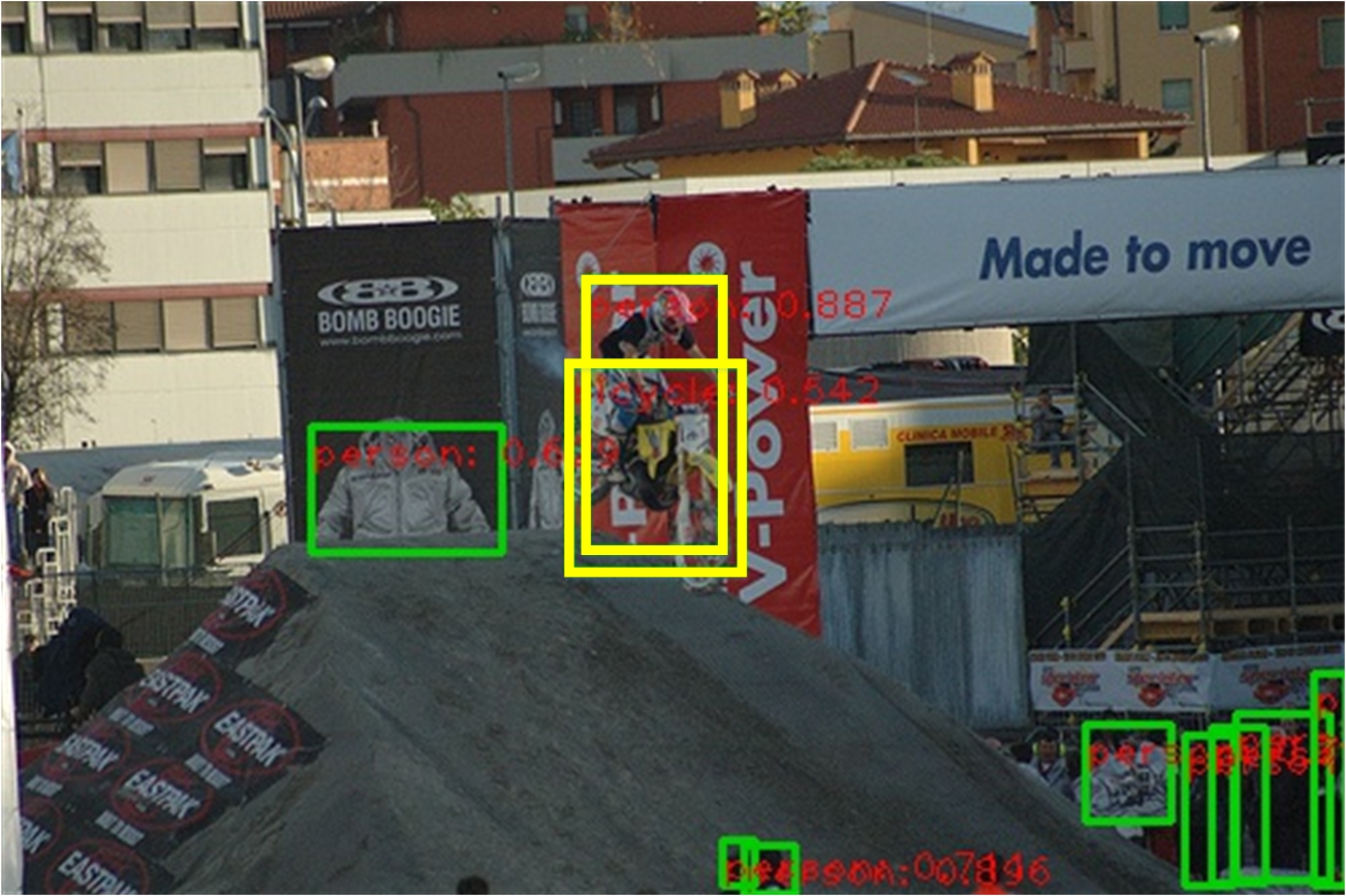}
		\caption{ }
	\end{subfigure}%
	\begin{subfigure}[b]{.33\linewidth}
		\centering
		\includegraphics[width=.9\textwidth]{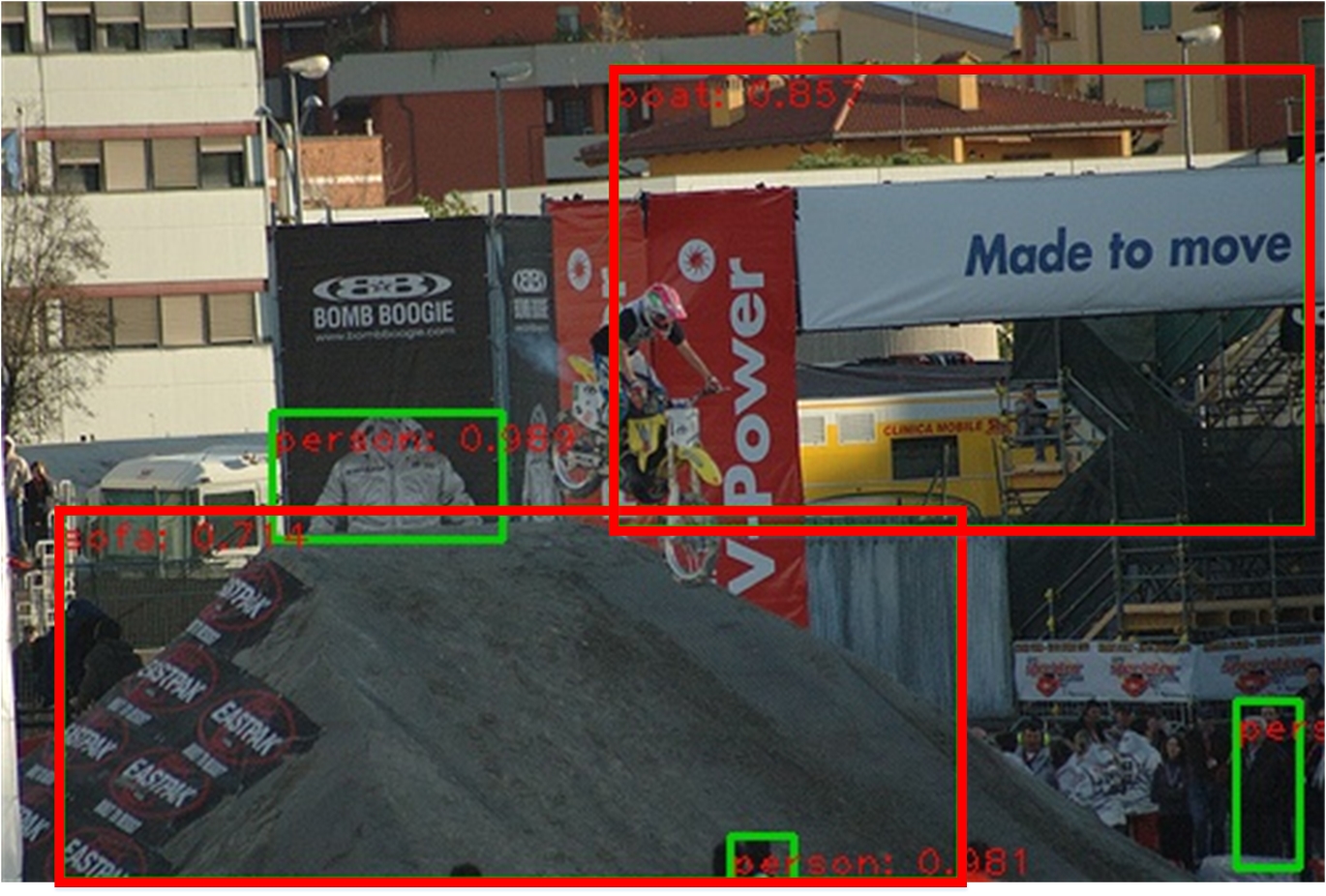}
		\caption{ }
	\end{subfigure}%
	\caption{The detection results of a) original student network, b) the KTAN network, and c) teacher network. The yellow bounding boxes in the middle column represent the increased correct detection results after the knowledge transfer and the red bounding box in left column represents the wrong detection in the original student network. (Best seen in color)}
\end{figure}

\subsection{Object detection}

Although, several works have successfully improved the ability of small networks on image classification problems, few of them attempt to explore the performance of their methods on other computer vision tasks, like object detection.

In this section, we set experiments to compare different methods, including KD \cite{hinton2015distilling}, FitNet \cite{romero2014fitnets}, DLN and our KTAN on object detection task. We evaluate them on PASCAL VOC 2007 dataset \cite{Everingham10}. On this dataset, we select the Faster-RCNN network as the object detection architecture, then load ResNet152 \cite{he2016deep} as teacher model and ResNet-50 \cite{he2016deep} as student model respectively. Following settings in \cite{girshick2015fast}, we train models on VOC 2007 trainval, and evaluate them on the test set with the mean Average Precision (mAP).

As shown in Table 2, our KTAN model achieves the best performance on PASCAL VOC 2017 dataset. Different from the classification task, object detection problems contain two optimizing targets, including predicting bounding boxes and classification results for all instances. In the Faster-RCNN method, it applies a region proposal network (RPN) to generate candidate bounding boxes from the output feature map of the last convolutional layer. Then, it maps candidate bounding boxes into feature map and classifies each bounding boxes. Therefore, a softmax output of a large model only contains the probability distribution of candidate bounding boxes. Differently, our KTAN model extracts the last convolutional layer’s feature maps of a large teacher model as shared knowledge, which contains the understanding of the large model overall input image. Besides, through the adversarial training process, our KTAN method can transfer more spatial information from a large model to a small one than DLN method.

Besides, we also provide some detection results of KTAN on PASCAL VOC 2007 test set. As shown in Figure 2, the first line represents the detection results of original student network, the last line represents teacher networks' results, and the middle line shows the prediction of our KTAN network. The yellow bounding boxes in the middle line represent the increased correct detection results after the knowledge transfer. Through our KTAN method, a simple Faster-RCNN model can generate better feature maps with the shared knowledge of a large Faster-RCNN model than the original one. Therefore, it can detect more correct bounding boxes than the original model. With a better feature map, the improved model can also remove the wrong detection bounding boxes, as shown in the fourth column. Besides, our KTAN model only learns an intermediate representation from a teacher network. Therefore, it can get rid of some wrong predictions from the teacher network.

\section{Conclusion}

In this paper, we propose a deep feature maps knowledge based adversarial knowledge transfer framework for various computer vision tasks, which is implemented in two sequential processes. For the knowledge extraction operation, we propose to transfer the intermediate representation of teacher to student network. Furthermore, a Teacher-to-Student layer is designed to adapt to different student neural network structures. Different from previous directly matching knowledge learning solution, considering that the most valuable information carried by the shared knowledge is the spatial structure and correlation between feature maps, we devise an adversarial training framework to teach the student network to understand the spatial information hidden in the shared knowledge. The results are quite encouraging, which further proves that our method indeed helps the student network learn from the generalization of teacher network and show better performance.

We believe the split understanding about the traditional KD would help to understand the contributions of previous works and also helpful for further research. In the future work, we aim to explore more powerful adversarial framework and pursue more applications of our KTAN methods, specifically on computer vision tasks, like video caption, video semantic understanding and so on.

\bibliography{aaai}

\begin{thebibliography}{}

\bibitem[\protect\citeauthoryear{Cheng \bgroup et al\mbox.\egroup
  }{2017}]{cheng2017survey}
Cheng, Y.; Wang, D.; Zhou, P.; and Zhang, T.
\newblock 2017.
\newblock A survey of model compression and acceleration for deep neural
  networks.
\newblock In {\em arXiv preprint arXiv:1710.09282}.

\bibitem[\protect\citeauthoryear{Courbariaux \bgroup et al\mbox.\egroup
  }{2016}]{courbariaux2016binarized}
Courbariaux, M.; Hubara, I.; Soudry, D.; El-Yaniv, R.; and Bengio, Y.
\newblock 2016.
\newblock Binarized neural networks: Training deep neural networks with weights
  and activations constrained to+ 1 or-1.
\newblock In {\em arXiv preprint arXiv:1602.02830}.

\bibitem[\protect\citeauthoryear{Everingham \bgroup et al\mbox.\egroup
  }{2010}]{Everingham10}
Everingham, M.; Van~Gool, L.; Williams, C. K.~I.; Winn, J.; and Zisserman, A.
\newblock 2010.
\newblock The pascal visual object classes (voc) challenge.
\newblock In {\em International Journal of Computer Vision}, volume~88,
  303--338.

\bibitem[\protect\citeauthoryear{Geifman and
  El-Yaniv}{2017}]{geifman2017selective}
Geifman, Y., and El-Yaniv, R.
\newblock 2017.
\newblock Selective classification for deep neural networks.
\newblock In {\em Advances in Neural Information Processing Systems},
  4885--4894.

\bibitem[\protect\citeauthoryear{Girshick}{2015}]{girshick2015fast}
Girshick, R.
\newblock 2015.
\newblock Fast r-cnn.
\newblock In {\em Proceedings of the IEEE International Conference on Computer
  Vision},  1440--1448.

\bibitem[\protect\citeauthoryear{Goodfellow \bgroup et al\mbox.\egroup
  }{2014}]{goodfellow2014generative}
Goodfellow, I.; Pouget-Abadie, J.; Mirza, M.; Xu, B.; Warde-Farley, D.; Ozair,
  S.; Courville, A.; and Bengio, Y.
\newblock 2014.
\newblock Generative adversarial nets.
\newblock In {\em Advances in Neural Information Processing Systems},
  2672--2680.

\bibitem[\protect\citeauthoryear{Han, Mao, and Dally}{2015}]{han2015deep}
Han, S.; Mao, H.; and Dally, W.~J.
\newblock 2015.
\newblock Deep compression: Compressing deep neural networks with pruning,
  trained quantization and huffman coding.
\newblock In {\em arXiv preprint arXiv:1510.00149}.

\bibitem[\protect\citeauthoryear{He \bgroup et al\mbox.\egroup
  }{2016a}]{he2016deep}
He, K.; Zhang, X.; Ren, S.; and Sun, J.
\newblock 2016a.
\newblock Deep residual learning for image recognition.
\newblock In {\em IEEE Conference on Computer Vision and Pattern Recognition},
  770--778.

\bibitem[\protect\citeauthoryear{He \bgroup et al\mbox.\egroup
  }{2016b}]{he2016identity}
He, K.; Zhang, X.; Ren, S.; and Sun, J.
\newblock 2016b.
\newblock Identity mappings in deep residual networks.
\newblock In {\em European Conference on Computer Vision},  630--645.

\bibitem[\protect\citeauthoryear{Hinton, Vinyals, and
  Dean}{2015}]{hinton2015distilling}
Hinton, G.; Vinyals, O.; and Dean, J.
\newblock 2015.
\newblock Distilling the knowledge in a neural network.
\newblock In {\em arXiv preprint arXiv:1503.02531}.

\bibitem[\protect\citeauthoryear{Howard \bgroup et al\mbox.\egroup
  }{2017}]{howard2017mobilenets}
Howard, A.~G.; Zhu, M.; Chen, B.; Kalenichenko, D.; Wang, W.; Weyand, T.;
  Andreetto, M.; and Adam, H.
\newblock 2017.
\newblock Mobilenets: Efficient convolutional neural networks for mobile vision
  applications.
\newblock In {\em arXiv preprint arXiv:1704.04861}.

\bibitem[\protect\citeauthoryear{Hu, Huang, and Schwing}{2017}]{hu2017maskrnn}
Hu, Y.-T.; Huang, J.-B.; and Schwing, A.
\newblock 2017.
\newblock Maskrnn: Instance level video object segmentation.
\newblock In {\em Advances in Neural Information Processing Systems},
  324--333.

\bibitem[\protect\citeauthoryear{Ioffe and Szegedy}{2015}]{ioffe2015batch}
Ioffe, S., and Szegedy, C.
\newblock 2015.
\newblock Batch normalization: Accelerating deep network training by reducing
  internal covariate shift.
\newblock In {\em International Conference on Machine Learning},  448--456.

\bibitem[\protect\citeauthoryear{Krizhevsky, Sutskever, and
  Hinton}{2012}]{krizhevsky2012imagenet}
Krizhevsky, A.; Sutskever, I.; and Hinton, G.~E.
\newblock 2012.
\newblock Imagenet classification with deep convolutional neural networks.
\newblock In {\em Advances in Neural Information Processing Systems},
  1097--1105.

\bibitem[\protect\citeauthoryear{Li \bgroup et al\mbox.\egroup
  }{2017}]{li2017pruning}
Li, H.; Kadav, A.; Durdanovic, I.; Samet, H.; and Graf, H.~P.
\newblock 2017.
\newblock Pruning filters for efficient convnets.
\newblock In {\em International Conference on Learning Representations}.

\bibitem[\protect\citeauthoryear{Liu, Liu, and Ma}{2017}]{liu2017weighted}
Liu, P.; Liu, W.; and Ma, H.
\newblock 2017.
\newblock Weighted sequence loss based spatial-temporal deep learning framework
  for human body orientation estimation.
\newblock In {\em IEEE International Conference on Multimedia and Expo},
  97--102.

\bibitem[\protect\citeauthoryear{Luo \bgroup et al\mbox.\egroup
  }{2016}]{luo2016face}
Luo, P.; Zhu, Z.; Liu, Z.; Wang, X.; Tang, X.; et~al.
\newblock 2016.
\newblock Face model compression by distilling knowledge from neurons.
\newblock In {\em AAAI Conference on Artificial Intelligence},  3560--3566.

\bibitem[\protect\citeauthoryear{Newell, Huang, and
  Deng}{2017}]{newell2017associative}
Newell, A.; Huang, Z.; and Deng, J.
\newblock 2017.
\newblock Associative embedding: End-to-end learning for joint detection and
  grouping.
\newblock In {\em Advances in Neural Information Processing Systems},
  2274--2284.

\bibitem[\protect\citeauthoryear{Romero \bgroup et al\mbox.\egroup
  }{2014}]{romero2014fitnets}
Romero, A.; Ballas, N.; Kahou, S.~E.; Chassang, A.; Gatta, C.; and Bengio, Y.
\newblock 2014.
\newblock Fitnets: Hints for thin deep nets.
\newblock In {\em International Conference on Learning Representations}.

\bibitem[\protect\citeauthoryear{Simonyan and
  Zisserman}{2014}]{simonyan2014very}
Simonyan, K., and Zisserman, A.
\newblock 2014.
\newblock Very deep convolutional networks for large-scale image recognition.
\newblock In {\em arXiv preprint arXiv:1409.1556}.

\bibitem[\protect\citeauthoryear{Wei \bgroup et al\mbox.\egroup
  }{2017}]{wei2017deep}
Wei, X.-S.; Zhang, C.-L.; Li, Y.; Xie, C.-W.; Wu, J.; Shen, C.; and Zhou, Z.-H.
\newblock 2017.
\newblock Deep descriptor transforming for image co-localization.
\newblock In {\em International Joint Conference on Artificial Intelligence}.

\bibitem[\protect\citeauthoryear{Zagoruyko and
  Komodakis}{2017}]{zagoruyko2016paying}
Zagoruyko, S., and Komodakis, N.
\newblock 2017.
\newblock Paying more attention to attention: Improving the performance of
  convolutional neural networks via attention transfer.
\newblock In {\em International Conference on Learning Representations}.

\bibitem[\protect\citeauthoryear{Zeiler and
  Fergus}{2014}]{zeiler2014visualizing}
Zeiler, M.~D., and Fergus, R.
\newblock 2014.
\newblock Visualizing and understanding convolutional networks.
\newblock In {\em European Conference on Computer Vision},  818--833.

\bibitem[\protect\citeauthoryear{Zhou \bgroup et al\mbox.\egroup
  }{2017}]{zhou2017incremental}
Zhou, A.; Yao, A.; Guo, Y.; Xu, L.; and Chen, Y.
\newblock 2017.
\newblock Incremental network quantization: Towards lossless cnns with
  low-precision weights.
\newblock In {\em International Conference on Learning Representations}.

\end{thebibliography}
\bibliographystyle{aaai}

\end{document}